\documentclass[twocolumn,epsfig]{article} 
\usepackage{graphicx} 

\newenvironment{maliste}%
{ \begin{list}%
        {$\bullet$}%
        {\setlength{\labelwidth}{30pt}%
         \setlength{\leftmargin}{20pt}%
         \setlength{\itemsep}{\parsep}}}%
{ \end{list} }

\newenvironment{changemargin}[2]{\begin{list}{}{%
\setlength{\topsep}{0pt}%
\setlength{\leftmargin}{0pt}%
\setlength{\rightmargin}{0pt}%
\setlength{\listparindent}{\parindent}%
\setlength{\itemindent}{\parindent}%
\setlength{\parsep}{0pt plus 1pt}%
\addtolength{\leftmargin}{#1}%
\addtolength{\rightmargin}{#2}%
}\item }{\end{list}}

\long\def\comment#1{}

\makeatletter
\def\@normalsize{\@setsize\normalsize{10pt}\xpt\@xpt
\abovedisplayskip 10pt plus2pt minus5pt\belowdisplayskip
\abovedisplayskip \abovedisplayshortskip \z@
plus3pt\belowdisplayshortskip 6pt plus3pt
minus3pt\let\@listi\@listI}

\def\subsize{\@setsize\subsize{12pt}\xipt\@xipt}
\def\section{\@startsection {section}{1}{\z@}{1.0ex plus
1ex minus .2ex}{.2ex plus .2ex}{\large\bf}}
\def\subsection{\@startsection
   {subsection}{2}{\z@}{.2ex plus 1ex} {.2ex plus .2ex}{\subsize\bf}}
\makeatother

\begin{document}

\date{}

\title{\huge \bf {Agent-Based Perception of an Environment \\ in an Emergency Situation }}

\author{Fahem Kebair, Fr\'ed\'eric Serin and Cyrille Bertelle
 \thanks{ Laboratoire d'Informatique, de Traitement de l'Information et des Syst\`emes,
 University of Le Havre, 25 rue Philippe Lebon, 76058, Le Havre Cedex, France. 
 Email: \{fahem.kebair, frederic.serin, cyrille.bertelle\}@univ-lehavre.fr
 }
}

\maketitle
\thispagestyle{empty}


{\hspace{1pc} {\it{\small Abstract}}{\bf{\small---We are interested in the problem of multiagent systems development for risk detecting and emergency response in an uncertain and partially 	perceived environment. The evaluation of the current situation passes by three stages inside the multiagent system. In a first time, the situation is represented in a dynamic way. The 	second step, consists to characterise the situation and finally, it is compared with other similar known situations. In this paper, we present an information modelling 
of an observed environment, that we have applied on the RoboCupRescue Simulation System. Information coming from the environment are formatted according to a taxonomy 
and using semantic features. The latter are defined thanks to a fine ontology of the domain and are managed by factual agents that aim to represent dynamically the current situation. 

 }}
 }


\section{Introduction}
\label{Introduction}

Recent catastrophic disasters have brought urgent needs for diverse technologies for disaster relief. Currently, there is an overwhelming need for better
information technology to help support the efficient and the effective management of the disaster management (also known as emergency response). In particular, 
actors and agencies need an assistance to help them to make a decision in a fashion time and to be able to coordinate their efforts in a flexible way in order 
to prevent further problems or effectively manage the aftermath of a disaster. Our project is situated in this context and consists to develop a generic 
Decision Support System (DSS), able to detect a risk in an uncertain and partially perceived environment and to prevent its evolution. The DSS kernel is a multiagent 
system with three layers, where each one has a specific role. The role of the lower layer, that we call the representation layer, is to represent 
the environment state and its evolution over the time. The environment is perceived as a whole of entities, directly or indirectly observable and of which states change 
permanently. These entities are modeled according to a taxonomic organisation and information that describe them are formatted according to a model of ``semantic features'', inspired by the memento design pattern rules \cite{gamma95}. Moreover, the system apprehends these information via software agents (called factual agents) and according to an ontology of the studied domain. The collaboration of these agents and their comparisons with each other, form dynamic agents clusters. The latter are compared by past known scenarios. The final object of the study is to permit to prevent the occur of a crisis situation and to provide an emergency management planning.
	
This modelling was elaborate starting from the game of Risk \cite{person05} and tested on the RoboCupRescue Simulation System (RCRSS) \cite{robocuprescue}. In this paper, we provide 
a modelling of information extracted from an observed environment in an emergency context. Inside the system, information are managed thanks to factual agents that interact 
by comparing each other. The modelling includes a definition of a taxonomy. The latter was applied to the RCRSS environment, for which we have defined an ontology of the domain. The 
structure of the paper is as follows: first we present the general architecture of the DSS and its internal kernel. Then, we define the taxonomic organisation of the perceived 
environment. After that, we present the RCRSS environment and the ontology of the domain. Finally, we present factual agents and some tests using graphic tools.

\section{Decision Support System}
\label{Decision Support System}

The role of the Decision Support System is quite wide. In general, the purpose is ``to improve the decision making ability of managers (and operating personnel) 
by allowing more or better decisions within the constraints of cognitive, time, and economic limits'' \cite{holspace96}. More specifically, the purposes of a DSS are:

\begin{maliste}
   \item Supplementing the decision maker,
   \item Allowing better intelligence, design, or choice,
   \item Facilitating problem solving,
   \item Providing aid for non structured decisions,
   \item Managing knowledge.
\end{maliste}
Decision makers need quick responses to events that take place at a continually increasing rate and they should incorporate an enormous amount of knowledge such as data, choices 
and consequences. Also they must have fast access to consistent, high-quality knowledge to compete \cite{kim05}. 

In our context, the DSS is used as an emergency management system, able to assist actors in urban disasters mitigation and to prevent them about potential future critical 
consequences. The system includes a body of knowledge which describes some aspects of the decision-maker's world and that comprises the ontology of the domain and past known scenarios. 

	\begin{figure}[h]
 	   \centering
 	   \includegraphics[width=8.5cm,height=7cm,angle=0]{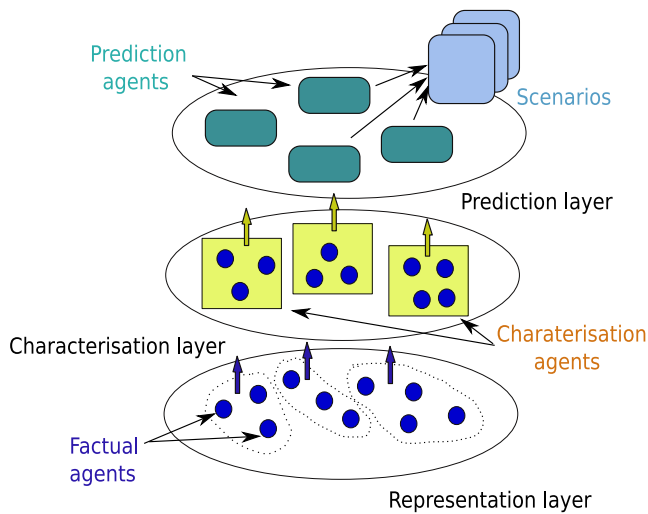}
 	   \caption{Kernel structure}
 	\end{figure}

The kernel of the DSS is a multiagent system with three layers. Agents of each layer have their own way of behaving and communicating.  

\textbf{Representation layer :}
This layer is composed by factual agents and has as essential aim to represent dynamically and in real time the information of the current situation. Each new entering information 
is dealt by a factual agent that intends to reflect a partial part of an observed situation. Agents interactions and more precisely, aggressions and mutual aids reinforce some agents 
and weaken some other. 

\textbf{Characterisation layer :}	
This layer has as aim to gather factual agents, emerged from the precedent layer, using clustering algorithms. We consider a cluster of agents, a group of which agents are close from 
dynamic and evolution manner point of view. The goal here, is to form dynamic structures, where each one is managed by a characterisation agent.
 
\textbf{Prediction layer :}
This layer is made up of prediction agents. Each one represents an observed scenario originally from the current situation. The task of the prediction agents is to compare 
their scenarios by past ones to provide a closed one of which result may be a potential consequence. This mechanism is based on the case base reasoning, 
the latter differs from a classic one by its ability to manage a dynamic and an incremental development.
	
\section{Taxonomic Organisation of the Studied Environment}
\label{Taxonomic Organisation of the Studied Environment}

Our perception of the environment focuses on two aspects: on the one hand, we observe the concrete objects of the world, the changes of their states and their interaction. On the other
hand, we observe the events and the actions that may be created naturally or artificially. We have defined therefore, three categories of objects (\figurename{ 2}): Concrete object, 
Action object and Message object.

\textbf{Concrete object :}
	Three types of concrete objects are distinguished. The first type is the Person object, which represents an actor of the environment. It is the only object that has the 
	ability to act and to interact and of which behaviour and state evolution are usually predictable. The second type is the Passive object. Two sub-categories are identified: 
	immobile objects as buildings and roads networks, and mobile objects like the means of transport. The observation of these objects is the simplest one, because they do not have any 
	behaviour. Their observation is reduced only to the description of their current state. The third type is the Mean object. It is created at a given time and for a particular 
	purpose. Its existence duration varies in time, according to the objective for which it is created. For example, a car is considered as a mean since it is driven by a driver, 
	otherwise, it is considered as an immobile object.

\textbf{Action object :}
   	This type is divided into activities and phenomena objects. Both are created at a given time and are limited temporally without a priory knowledge of the bounds. Phenomena are 
   	unpredictable events that start at a given time. Their observation is the most complex because of their uncertainties and their rapid evolutions. Activities are the actions 
   	sequences performed by actors. Generally, they are ordered and emitted for a particular purpose.
   	
\textbf{Message :}
   	Messages represent the interactions between persons and more precisely the information flows exchanged between the actors. The impact of a message is not easily measurable and it 
   	depends on its sender, its receiver and its performative. If the message is stored, its impact will be deferred.
   	
   	The observation of an object in the environment may concern a persistent, temporary or punctual state. A persistent state can become invalid following a rupture. For example, a 
   	building with ten floors is a persistent state as long as is not destroyed. A blocked road is also a persistent state until it will be unblocked. However, a fire is a temporary state, 
   	because it is foreseeable that it will cease, fault of combustible. Finally, a punctual state is immediate and instantaneous, like sending a message.

   	\begin{figure}[h]
	\hspace{-1cm}
 	   \includegraphics[width=9.5cm,height=4cm,angle=0]{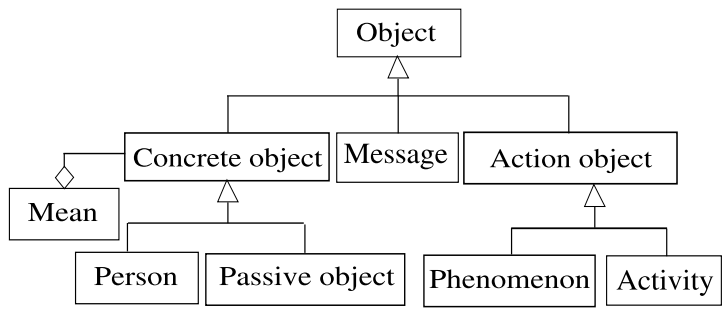} 
 	   \caption{Taxonomy of the observed environment}
 	\end{figure}
 	
\section{Formalisation of Information: Application on the RoboCupRescue Simulation System} 
\label{Formalisation of Information: Application on the RoboCupRescue Simulation System}	

\subsection{RoboCupRescue Simulation Project}
\label{RoboCupRescue Simulation Project}

RoboCupRescue (RCR) is an annual international competition within the framework of the RoboCup \cite{robocup}. This project intends to promote research and 
development in the disaster rescue domain by creating a standard simulator and forum for researchers and participators. RoboCupRescue project intends to simulate a urban disaster 
caused by an earthquake. The simulation disaster integrates various aspects of disasters. These includes, fire, housing and building damages, disruption of roads, electricity, water 
supply, gas, and other infrastructures, movements of refuges, status of victims, hospital operations, etc. RCRSS is composed by several distributed modules: a kernel, a geographic 
information system, simulators (fire, traffic and collapse simulators), a viewer and an RCR agents module. We are interested in our work in the RCR agents module. The work 
consists in designing rescue teams that have as mission to save civilians and mitigate disaster consequences. The final goal is to set up a strategy planning that permits 
teams coordination.
	
The picture \figurename{ 3} shows the hierarchy classes of the RCR disaster space. Each object in the world has properties such as its position, its shape ans its state. We distinguish 
two main objects categories: moving objects and motionless objects. First ones represent actors of the disaster world and they are modelled by Person object in our taxonomy. The second 
category consists of both buildings and networks roads and they are modelled by Passive object in the taxonomy.
	
Seven RCR agents types exist in the RCR simulation world: three platoon agents which are fire brigade, police force and ambulance team, three center agents which are fire 
station, police office and ambulance center and civilian agents. We will not develop the behaviours of the latter, because they are simulated independently with the other 
simulators. Each RCR agent has a partial knowledge of the whole environment state. This knowledge is updated thanks to two capacities: visual and auditory 
capacities. These capacities permit agents to receive information send by the kernel of the simulator each cycle (one second in the simulation which represents one minute in the 
reality). Agent centers represent in reality persons inside, so they can only see their surrounding area of the world and exchange messages with other agents. The role of these centers 
is to coordinate the communication between the three agents types. Platoon agents have more capacities, they can act by performing seven different actions: rescue, load and unload 
actions for ambulance team agents, extinguish for fire brigade agents, clear for police force agents and move for all agents.

\begin{figure}[h]
\centering
\includegraphics[width=9cm,height=10cm,angle=0]{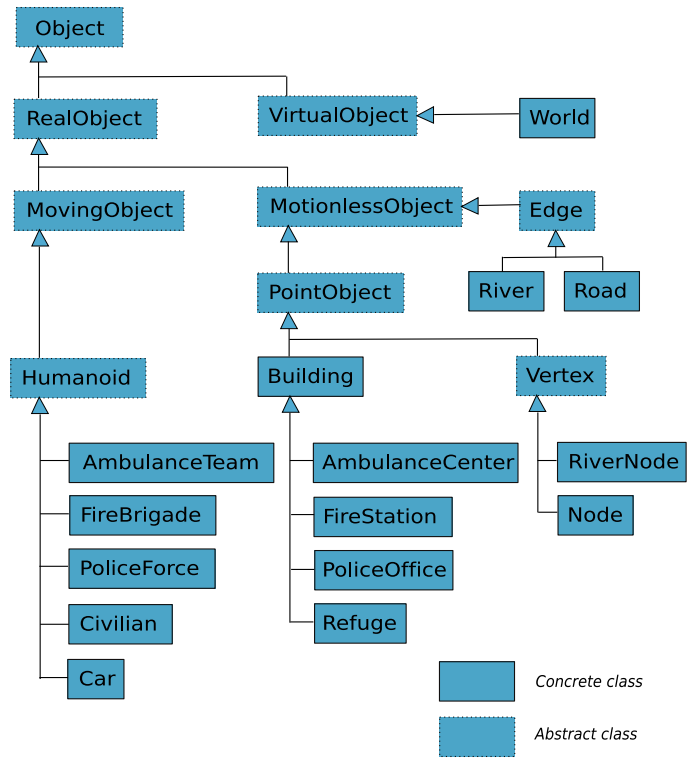}
\caption{Class hierarchy of the RCR objects in the disaster space}
\end{figure}

\subsection{Ontology of the Domain}
The definition of the ontology of the domain is the result of the taxonomy application on the RCR simulation environment. The determination of the concepts is based on the object 
modelling of the RCR environment and respects the taxonomic organisation. The next picture (\figurename{ 4}) shows the ontology that we have implemented using prot\'eg\'e \cite{protégé}. 

\begin{figure}[h]
\centering
\begin{changemargin}{-1cm}{0cm}
\includegraphics[width=10cm,height=7cm,angle=0]{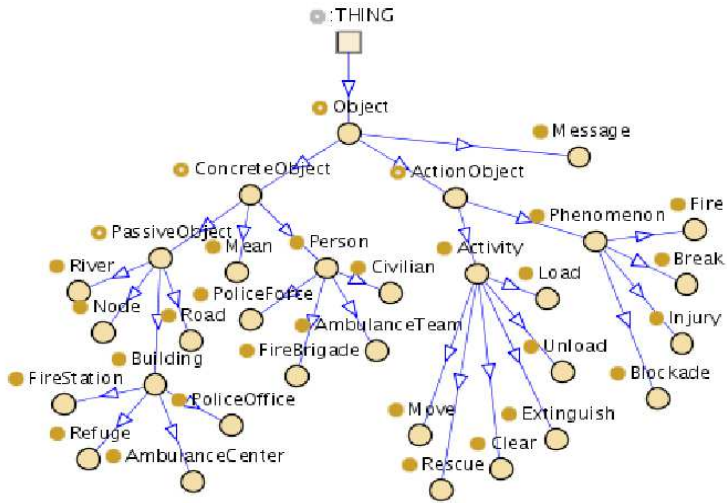}
\caption{Ontology of the RoboCupRescue environment}
\end{changemargin}
\end{figure}

The abstract class Object is situated on the top level of the classes hierarchy. Each object of the environment has a type and is localised in time and space. We have assigned 
therefore to Object class a type, a time and a localisation attributes. In the second level, three classes inherit the Object class. Two abstract classes: ActionObject and 
ConcreteObject, and a concrete class Message. 

ActionObject class is the superclass of Phenomenon and Activity classes. The first one is the superclass of Fire, Break, Injury and Blockade classes and has an additional attribute 
intensity. The latter represents the intensity and the progression degree of the phenomenon. For example, a fire may have the following intensities: starting, strongly and 
extremely$\_$strongly. Activity class is the superclass of Load, Rescue, Unload, Extinguish, Move and Clear which are the RCR agents actions defined above. This class has two 
additional attributes: actor and target. Actor attribute takes as value an RCR agent name and target attribute has as value a Concrete object name that may be: a building, a road, 
a civilian, etc.
	
ConcreteObject class is the superclass of the concrete classes: Person, PassiveObject and Mean classes. Person class has three additional attributes: buriedness, damage and hitPoint. 
The first one shows how much a person is buried in the collapse buildings. The second one shows the necessity of medical treatment. The last one shows the health level, a person in 
good health has a hitPoint = 10000, and 0 when his is dead. PassiveObject and Mean classes has only the inherited attributes. 

Finally, Message class is a concrete class and has two additional attributes: receiver and sender. In RoboCupRescue, a message content has the following format: ``action\_name 
object\_name''. For example, ``clear road$\#$ID'', ``extinguish building$\#$ID'', or ``rescue civilian$\#$ID'', etc. The localisation attribute means therefore more precisely, the 
localisation of the target object in the message content as the road$\#$ID, building$\#$ID and civilian$\#$ID in these examples.

\subsection{Semantic Features}
Information coming from the environment are written in the form of semantic features. The latter will be managed thereafter by factual agents in the representation layer. The idea to 
use a semantic feature is inspired from the memento design pattern and consists to store information, that describe the internal state of an observed object originally from the 
taxonomy. The structure of a semantic feature is generic and composed by a key and a set of couples $<$qualifier,value$>$. The key is defined from the taxonomy and the qualifiers 
are defined from the ontology. 
	
In RoboCupRescue, information are sent by RCR agents each cycle and may be visual or auditive information. The system treats these data in order to extract the important ones. For 
example, an RCR agent who sends an information describing an intact building will not be taken into account. However, an information about a burning building is interesting, 
the system interprets it and creates thereafter new semantic features, related to objects defined by the taxonomy. As example, a Building$\#$14 has a property ``fieryness = 25'', this 
means that a fire has just started in this building. The system creates therefore, a semantic feature: (Phenomenon$\#$14, type, fire, intensity, starting, localisation, 20$|$25, time, 
7). This semantic feature is related to a phenomenon object, that means a fire is located in 20$|$25 coordinates at the seventh cycle of the simulation. In the case of an auditive 
information, the system creates semantic features according to messages contents. For example, an RCR agent sends a message "clear road$\#$15". From this message, a semantic feature 
(Phenomenon$\#$22, type, blockade, intensity, unknown, localisation, 30$|$40, time, 11) is created. This semantic feature is related to a blockade phenomenon, a priory we do not know 
the intensity of the blockade, but we can determine the coordinates of the blocked road from the world map, using its identifier (15). Thus, by treating the messages and the visual 
information sent by RCR agents, the system gathers the partial knowledges of these agents to build a global knowledge that can provide a clearer idea about the situation.
	
Semantic features are related with each other, that means they have a semantic dependencies. We defined therefore proximity measures in order to compare between them. The proximity 
value is comprised between [-1,1]. Two semantic features are opposite in their subjects if the proximity measure is negative, they are closed if it is positive and independent if it 
equals zero. More the proximity is near to 1 (-1), more the two semantic features are closed (opposite). We distinguish three types of proximities: a semantic proximity which is 
determined thanks to the ontology, a spatial and a time proximities that are related to specific scales. As example, a break and a block are closed semantically, because if a building 
is broken, the nearest road will be certainly blocked. Moreover, to give more precision to this confrontation, we compare the localisations and the times of observation of the two 
events. If they are distant, we consider the two events are independent, and inversely. 

\section{Factual Agents of the Representation Layer}
   \subsection{Structure and Role}
The representation layer is composed by factual agents. Each agent aims to represent a partial part of the observed situation, thanks to the semantic feature that it 
carries. 

	\begin{figure}[h]
 	   \centering
 	   \includegraphics[width=4cm,height=4cm,angle=0]{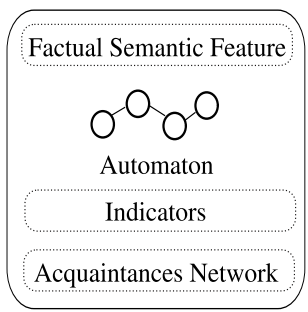}
 	   \caption{Internal structure of a factual agent}
 	\end{figure}
 	
Factual agent is a reactive and proactive agent \cite{wood02}. Its reactivity is ensured by a generic internal behavioural automaton of Augmented Transition Network (ATN) 
type \cite{woods70}. This automaton is composed by four states \cite{cardon04}: initialisation, deliberation, decision and action. ATN Transitions are stamped by a set of conditions 
and a sequence of actions. Conditions represent thresholds, defined according to three internal indicators of the agent, which are: PseudoPosition (PP), PseudoSpeed (PS) and 
PseudoAcceleration (PA). The agent has two other indicators: a satisfaction indicator and a constancy indicator, which represent respectively the satisfaction degree of the agent 
about its progression and its stability in its ATN. The definition of these indicators allow the factual agent to progress in its ATN, this characteristic ensures the proactivity of 
the agent, of which purpose is to achieve the most important state, that is the action state. In addition, the factual agent is a social agent. It interacts with the other agents in 
order to form a coalition with other ones, this permits it to acquire more force and power. The agent can also be attacked by other ones, with which it is opposite semantically. The 
list of opposites agents and closes agents is stored in an acquaintances network, which is constructed and updated dynamically.
	
   \subsection{Tests and Graphic Tools}
  
We started to make tests on a part of the ontology. We localised our tests especially on the detection of the different events signalled by the RCR agents and the actions that they perform. 
	
	\begin{figure}[h]
 	   \centering
 	   \includegraphics[width=9cm,height=7.5cm,angle=0]{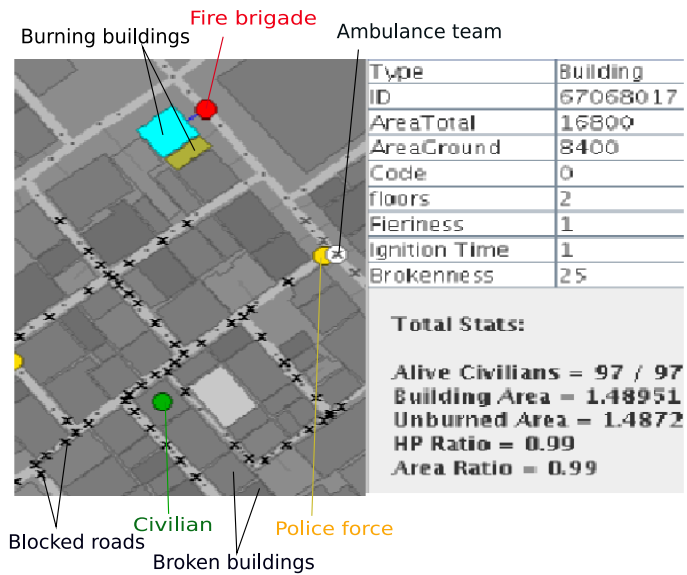}
 	   \caption{View of the RCR disaster space}
 	\end{figure}
 		
We have designed some graphic tools in order to follow and study the evolution of the factual agents. 
	
	\begin{figure}
	\begin{center}
	\includegraphics[width=8cm,height=8cm,angle=0]{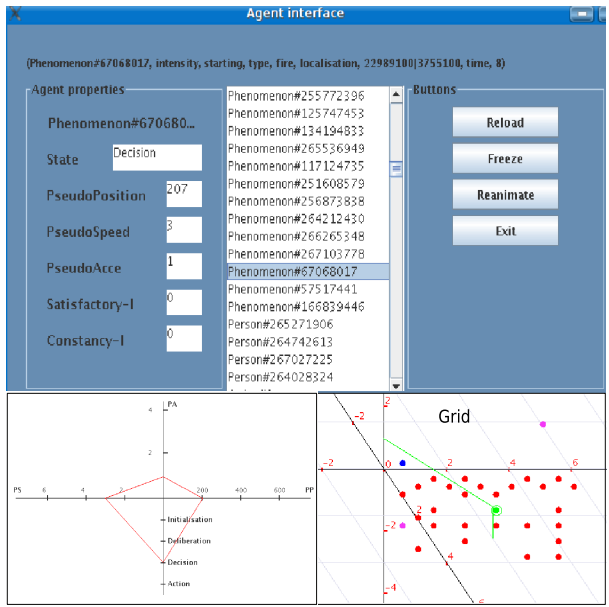}
	\caption{Graphic tools to visualise factual agents}
	\end{center}
	\end{figure}	
	
The graphic tool is composed by a grid that shows in real time points flow representing factual agents. Agents are projected on three axis: 
PP, PS and PA. Factual agents progress extremely quickly, so it is too hard to follow their evolution. We have created therefore, an interactive interface (agent interface). This 
interface has two essential functionalities. The first one permits to select a given factual agent and to show all its information: its semantic feature, its current 
state and its current indicators values. The second one permits to freeze all the factual agents at a given time and to reanimate them thereafter. This allows us to obtain an 
instantaneous view of all the agents during their evolution and to study consequently, information about any agent.
	
Picture \figurename{ 6} shows an instantaneous image of the current situation of the RCRSS disaster space in the eighth cycle of the simulation. Information shown in the table, in 
the right, are related to the blue building, that is burning. A new factual agent, carrying the semantic feature (Phenomenon\#67068017, type, fire, intensity, starting, localisation 
22989100$|$3755100, time, 8), is created and updated according to information sent by the fire brigade agent, situated just near to the building. This factual agent is represented by 
the green ellipse in the grid and has as coordinates (PP=207,PS=3,PA=1). In the agent interface, we can see all information about this agent, notably, its indicators and its state which is the 
decision state. We note, that all indicators are strictly positive and the agent is in advanced state in its ATN. This means the agent has acquired importance and the event that it 
represents is more and more significant. This evolution is the result of information sent by the fire brigade agent and the interaction of the factual agent with other 
factual agents. The latter carry other related information, that can be messages announcing the fire, or actions performed to extinguish it.

 \section{Conclusion}
This paper has presented an information modelling of a perceived environment in an emergency context. This modelling is used to represent the evolution of the current situation 
thanks to factual agents. Our final goal is to build a generic mutliagent system that intends to detect a risk an to deal with it. Information entering to the system are structured in 
the form of semantic features. The latter are defined thanks to a taxonomic organisation and to an ontology related to the domain. We choose the RCRSS as application to test this 
modelling. We have implemented therefore, the ontology of the studied domain and started the representation of the RCRSS disaster space state, using a part of the ontology. This test 
allowed us to study the behaviour of factual agents and especially ATN thresholds and proximity measures that are very dependant on the application and that require more control 
of the environment in order to validate them. Our future work consists in finishing both, the implementation of the ontology and the representation layer that are still under 
realisation. We have the intention thereafter, to connect this layer with the characterisation layer in order to test some observed scenarios of which definition constitutes also a 
future subject of study.

FAHEM KEBAIR is a PhD student in computer science since 2006. His application domain concerns Agent-Based Software Engineering. 

FREDERIC SERIN received his Ph.D. degree in Object-Oriented Simulation in 1996 from University of Rouen. Dr. Serin is currently Associate Professor at University of Le Havre. His current application domain now concerns Agent-Based Software Engineering.

CYRILLE BERTELLE is professor in computer science in Le Havre University and develops research activities in complex systems modeling. He is one of the co-directors of a 
research laboratories amalgamation which promotes the Sciences and Technologies in Information and Communication over the Haute-Normandie in France.


\end{document}